\documentclass[11pt]{article}
\usepackage[hyperref]{ccl2025-en}
\usepackage{times}
\usepackage{url}
\usepackage{latexsym}
\usepackage{fancyhdr}
\usepackage{amsmath}
\usepackage{CJKutf8}
\usepackage{booktabs}
\usepackage{graphicx}
\usepackage{hyperref}

\title{System Report for CCL25-Eval Task 6: Chinese  Essay Rhetoric Recognition Using LoRA, In-context Learning and Model Ensemble}

\author{Yuxuan Lai\thanks{~~Corresponding author}, ~~Xiajing Wang \and Chen Zheng \\
        \textsuperscript{1}The Open University of China, Beijing, China \\ \textsuperscript{2}Engineering Research Center of Integration and Application of Digital Learning Technology, \\ Ministry of Education, Beijing, China \\ {\tt erutan@pku.org.cn}}

\date{}

\begin{document}
\begin{CJK*}{UTF8}{gbsn}
\maketitle
\begin{abstract}
Rhetoric recognition is a critical component in automated essay scoring. By identifying rhetorical elements in student writing, AI systems can better assess linguistic and higher-order thinking skills, making it an essential task in the area of AI for education.
In this paper, we leverage Large Language Models (LLMs) for the Chinese rhetoric recognition task.
Specifically, 
we explore Low-Rank Adaptation (LoRA) based fine-tuning and in-context learning to integrate rhetoric knowledge into LLMs.
We formulate the outputs as JSON to obtain structural outputs and translate keys to Chinese.
To further enhance the performance, we also investigate several model ensemble methods.
Our method achieves the best performance on all three tracks of CCL 2025 Chinese essay rhetoric recognition  evaluation task, winning the first prize.
\englishkeywords{Large language model \and In-context learning \and LoRA \and Rhetoric recognition}
\end{abstract}

\section{Introduction}
\label{intro}

\cclfootnote{
    %
    %
    \hspace{-0.65cm}  
    \textcopyright 2025 China National Conference on Computational Linguistics

    \noindent Published under Creative Commons Attribution 4.0 International License
}

In written discourse, rhetorical devices such as metaphors, similes, and parallelism play a crucial role in elevating expressiveness and conveying subtle meaning. Recognizing these rhetorical elements is indispensable for achieving in-depth text comprehension and evaluation. 
In recent years, the rhetoric recognition task has attracted increasing attention, bringing renewed vitality
~\cite{chen2021jointly,zhu2022configure,li2022new,firsich2024can,kuhn2024computational,liu2024cerd,nuowei-etal-2024-chinese}. 
Particularly in the context of automated essay scoring, the ability to identify and interpret rhetoric contributes significantly to assessing surface-level language proficiency as well as the sophistication and effectiveness of students’ writing.

The recent advancement of Large Language Models (LLMs) has brought remarkable progress to various natural language processing tasks~\cite{achiam2023gpt,qwen3}. However, despite these advances, the rhetoric recognition task remains particularly challenging. 
First, as an information extraction task, it requires models to identify the applied rhetorical types in the input content as well as extracting rhetoric components, like tenor and vehicle in a metaphor. 
This structural output format deviates the free-form generative nature of LLMs. 
Second, identifying rhetorical devices often demands specialized linguistic and contextual knowledge, posing a significant challenge in effectively incorporating such domain-specific knowledge into LLMs.

In this paper, we focus on leveraging LLMs to address the Chinese rhetoric recognition task. To bridge the gap between the free-form generation style of LLMs and the structured output, we adopt a JSON-based output format, following recent approaches that demonstrate LLMs are effective in generating JSON outputs~\cite{lee2023making,shorten2024structuredrag}. Furthermore, we translate the keys in the JSON schema into Chinese, making the structure more aligned with natural language and thus facilitating the models better reasoning and generation.

To effectively integrate domain-specific knowledge into LLMs for enhanced performance on the rhetoric recognition task, we explore two data-efficient methods: Low-Rank Adaptation (LoRA) based fine-tuning~\cite{hu2022lora} for open-source LLMs and in-context learning ~\cite{brown2020language} for close-source LLMs. 
Besides, we also investigate the combination of LoRA and in-context learning, i.e., augmenting predictions of LLMs after LoRA with additional in-context learning examples.

We further explore several model ensemble strategies to boost overall performance. 
For rhetoric type classification tasks, we employ a linear-weighted ensemble method.
In the case of rhetoric component extraction, we implement a simple yet effective fallback mechanism to handle scenarios where LLMs fail to generate valid outputs, generally due to LLMs' content safety restrictions or JSON parsing errors.

We conduct experiments on CCL 2025 Chinese essay rhetoric recognition evaluation task, which includes three tracks, 
i.e., fine-grained form-level rhetoric type classification, fine-grained content-level rhetoric type classification, and rhetorical component extraction.
Our method achieves scores of 47.18, 54.03, and 39.94 on the three tracks, respectively. 
Our method ranks first, and outperforms the second team by a large margin, demonstrating the effectiveness of proposed approach.



\section{Task Formulation}

In CCL 2025 rhetoric recognition   evaluation task, given a paragraph in Chinese, $P = \{S_1, S_2, \ldots, S_n\}$, where each $S_i$ denotes the $i$-th sentence in the paragraph.
The goal is to detect all sentence subsets $S_r \subseteq P$, identify their fine-grained rhetoric type at the form- and content-level, and extract rhetorical components like tenor and vehicle in a metaphor.
An example is provided in Table~\ref{table-example}, and the label sets are listed in Appendix~\ref{appendix:dataset}.


\begin{table}[htbp]
\centering
\label{table-example}
\begin{tabular}{p{\textwidth}}
\toprule
\textbf{Input}:
Sentence-1.庄稼汉们站在地头，望着这片黄澄澄像狗尾巴的稻谷。Sentence-2.他感到十分开心。\\
\textbf{Track-1} : Sentence-id: $\left\{1\right\}$.~~~~ form-level fine-grained type: 比喻-明喻(Metaphor-Simile) \\
\textbf{Track-2}: Sentence-id: $\left\{1\right\}$.~~~~~ content-level fine-grained type: 比喻-实在物比喻(Metaphor-Concrete) \\
\textbf{Track-3}: Sentence-id: $\left\{1\right\}$. ~~~~~ 本体(tenor)-稻谷(swathes of rice) ~~~~~  喻体(vehicle)-狗尾巴(tails of dogs) ~~~~~  喻词(comparator)-像(like) \\
\bottomrule
\end{tabular}
\caption{Example of expected outputs for Track-1, Track-2, and Track-3.}
\end{table}

\section{Methods}

\begin{figure}[t!]
    \centering
    \includegraphics[width=0.95\textwidth]{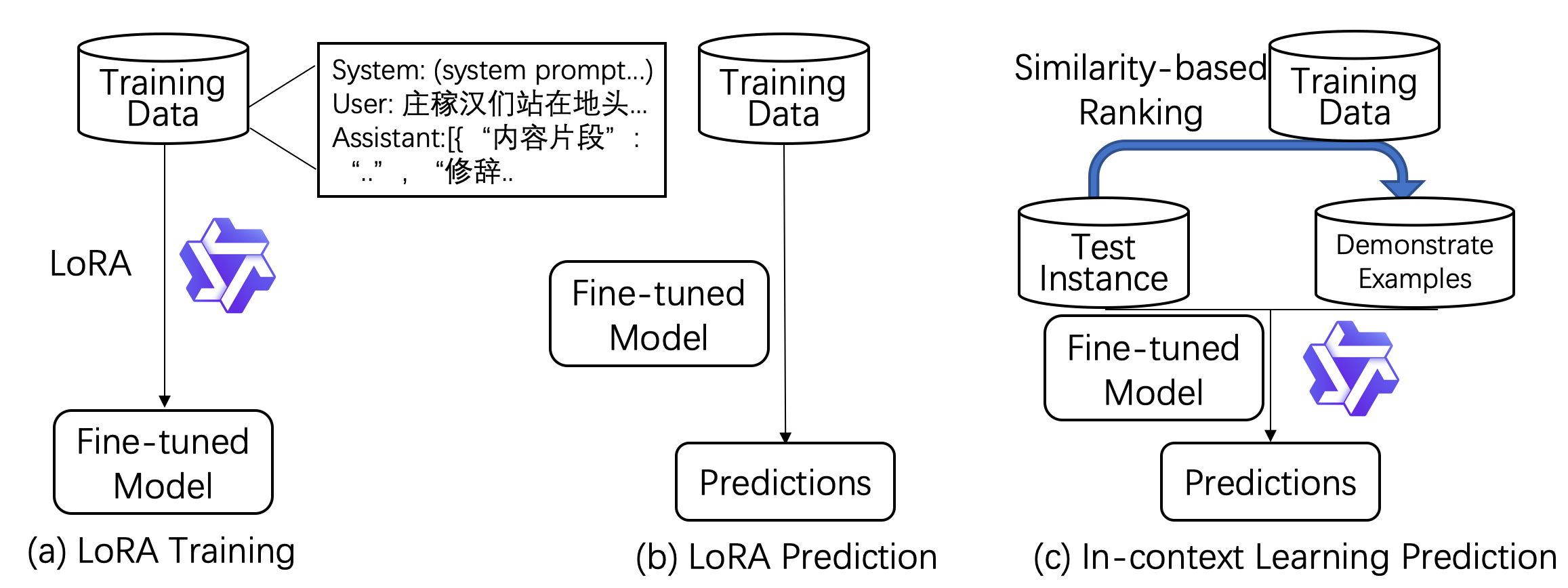} 
    \caption{An illustration of our methods.} 
    \label{fig:p1} 
\end{figure}

Followed recent works~\cite{jinwang-etal-2024-essay,myxiuci}, we investigate LoRA and in-context learning methods, and jointly train the three tasks. The framework is shown in Figure~\ref{fig:p1}.



\subsection{JSON-format Output}
\label{subsec:json}

As demonstrated in Table~\ref{tab:exampleJSON},
we organize the outputs in JSON format, and translate all keys into Chinese.
We avoid using numerical representations in the output, like Sentence IDs and conjunctionBeginIdx in the original annotation.
For some fields that may have multiple strings, such as \textit{tenor}, we use the special symbol \& to concatenate discontinuous rhetorical components into a single string.
These strategies aim at aligning the JSON content more closely with natural language.
We argue that making the JSON format more closer to natural language can better leverage LLMs' generation ability.
We recover these information from LLMs' predictions during post-processing (see \S\ref{subsec:postprocess}). 

Notice that the JSON-formatted output is used for joint training of the three tracks. Annotations from these tracks are integrated using heuristic rules, with only a minor portion (less than 10\%) requiring manual intervention. 
We argue that under joint learning manner, LLMs can integrate information from three tracks, thus making better decisions. 


\begin{table}[htbp]
\centering
\label{tab:exampleJSON}
\begin{tabular}{p{\textwidth}}
\toprule
\textbf{Inputs} : 庄稼汉们站在地头，望着这片黄澄澄像狗尾巴的稻谷。他感到十分开心。 \\
\textbf{Original Annotation}: Sentence ID: 1, ~ rhetoric: 比喻, ~ form: 明喻, ~ content: 实在物, ~ conjunction: [像], ~ conjunctionBeginIdx:[16], ~ conjunctionEndIdx:[16], ~
tenor:[稻谷], ~ 
tenorBeginIdx:[21], ~ 
tenorEndIdx:[22], ~ 
vehicle:[狗尾巴], ~ 
vehicleBeginIdx:[17], ~ 
vehicleEndIdx:[19] \\ 
\textbf{JSON output + translation}: 
[\{``内容片段": ``庄稼汉们站在地头，望着这片黄澄澄像狗尾巴的稻谷。", ``修辞手法": ``比喻", ``形式上的细粒度修辞分类": ``明喻", ``内容上的细粒度修辞分类": ``实在物", ``喻词": ``像", ``本体": ``稻谷", ``喻体": ``狗尾巴"\}]\\
\bottomrule
\end{tabular}
\caption{An Example for the JSON-format output.}
\end{table}

\subsection{Training and Testing Format for LoRA}
\label{subsec:formatlora}
The training and testing data for LoRA fine-tuning are formatted as single-turn dialogues that include a system prompt. The overall template is illustrated as follows. 
Within the system prompt (see Appendix~\ref{appendix:system_prompt}), we provide LLMs with explicit definitions of various rhetorical types and components, along with illustrative examples. 
The user input consists of the raw document text without any modifications. 
The model's target output during training is the well-structured JSON-format response described in \S~\ref{subsec:json}.

\begin{table}[htbp]
\centering
\begin{tabular}{p{\textwidth}}
\{``messages": [\{``role": ``system", ``content": \texttt{SYSTEM PROMPT}\}, \{``role": ``user", ``content": \texttt{INPUT DOCUMENT}\}, \{``role": ``assistant", ``content": \texttt{JSON OUTPUT}\}]\} \\
\end{tabular}
\end{table}

\subsection{Example Demonstration for In-context Learning}
\label{subsec:example}

Following prior works~\cite{luocontext,myxiuci}, we investigate a similarity-based method for selecting and ordering in-context examples. 
Given a test instance, we select the top- $k$ most similar instances from the training data to serve as the demonstrations. 
The more similar a training instance is to the test instance, the closer it is placed to the test instance.

Specifically, the similarity between instances is computed as the cosine similarity between their document embeddings. 
The input and output formats of demonstrated examples are consistent with those employed in LoRA fine-tuning (\S\ref{subsec:formatlora}). 
Our pilot study reveals that more complex query designs for test instances generally lead to a degradation in performance.
Consequently, we opt to simply use the input document of the test instance as the query for LLMs. 
The organization of the query for in-context learning is illustrated as follows:

\begin{table}[htbp]
\centering
\begin{tabular}{p{\textwidth}}
[\{``role": ``system", ``content": \texttt{SYSTEM PROMPT}\}, \{``role": ``user", ``content": \texttt{INPUT DOCUMENT$_1$}\}, \{``role": ``assistant", ``content": \texttt{JSON OUTPUT$_1$}\}, ......, \{``role": ``assistant", ``content": \texttt{JSON OUTPUT$_k$}\}, \{``role": ``user", ``content": \texttt{INPUT DOCUMENT$_{\texttt{TEST}}$}\}] \\
\end{tabular}
\end{table}

Considering that the coarse-grained rhetoric labels are inconsistent across the three tracks (see Appendix~\ref{appendix:dataset}), we also explore a strategy that separates the shared labels from the track-specific ones. 
Specifically, the four coarse-grained rhetoric types (metaphor, personification, hyperbole, and parallelism), which are present in all three tracks, are trained jointly, while the remaining four are trained separately. This approach is denoted as \texttt{SEPA}, and we denote the approach that jointly trains all rhetoric types as \texttt{JOIN}. 
We argue that this strategy brings two key advantages: it shortens the context length during in-context learning, and enhances the consistency of the output format across all rhetoric types.


\subsection{Post-processing}
\label{subsec:postprocess}




We design a series of post-processing strategies to adapt the raw outputs generated by LLMs into evaluation-ready predictions.
First, we employ Python’s built-in \textit{eval} function to parse the model outputs. For parsing failures, we assume that the model does not predict any rhetorical device. 
To determine the scope of each predicted rhetoric device, a sentence \textit{S} in the input document is considered as part of a rhetorical structure if the length of the longest common subsequence between the predicted \textit{内容片段} (see Table~\ref{tab:exampleJSON}) and \textit{S} is at least $\theta = max\left(5, length\left(S\right) * 0.6\right)$.
If the coarse-grained rhetoric type is missing from the output, the corresponding prediction is discarded. When the fine-grained rhetoric type is absent, we assign the most frequently occurring fine-grained type associated with the predicted coarse-grained type from the training data as a fallback.
For rhetorical component extraction, we extract the longest common substring between the predicted component and the selected sentences,
and simultaneously derive its start and end positions accordingly.

\subsection{Model Ensemble}
\label{subsec:ensemble}




For the rhetoric type classification task, we adopt a linear-weighted ensemble method. 
Let $\{M_1, M_2, \ldots, M_m\}$ denote the set of $m$ base models. The ensemble score indicating whether a sentence subset $S_r \subseteq P$ contains a specific rhetoric type $R_t$, is computed as follows:
$$
\mathrm{Score}(S_r, R_t) = \sum_{i=1}^m \left( \mathrm{pos}_i \cdot {I}_i(R_t) - \mathrm{neg}_i \cdot (1 - {I}_i(R_t)) \right)
$$
where ${I}_i(R_t)$ is an indicator function that equals 1 if model $M_i$ predicts that $S_r$ contains rhetoric $R_t$, and 0 otherwise. $\mathrm{pos}_i$ and $\mathrm{neg}_i$ represent the positive and negative weights assigned to model $M_i$, respectively. In our implementation, we simply set $\mathrm{neg}_i = \mathrm{pos}_i$ for all models.
A prediction is made if the $\mathrm{Score}(S_r, R_t)$ exceeds a predefined threshold $\theta_e$.


In the case of rhetoric component extraction, where model ensemble cannot be directly applied, we implement a fallback mechanism. 
Specifically, given a sequence of models $\{M_1, M_2, \ldots, M_m\}$, the prediction from model $M_i$ is utilized if and only if all preceding models $M_1, \ldots, M_{i-1}$ have failed to produce valid outputs.
This is generally due to content safety restrictions of LLMs or JSON parsing errors.

\section{Experiment}

\subsection{Experimental Setups}
\label{subsec:experimentalsetups}

We conduct experiments on  CCL 2025 rhetoric recognition   evaluation dataset. 
The training set comprises 50 instances, while the blind test set contains 37,459 instances.
Evaluation metrics involve the F1 scores for coarse- / fine-grained rhetoric classification, the F1 score for extraction of rhetoric components, and the intersection over union (IoU) score of sentence IDs.
For more details, see Appendix~\ref{appendix:dataset}.


We use Qwen 2.5 72B \cite{qwen2.5} as backend for LoRA and trained for 24 epochs (marked as $M_{\text{LoRA}}$). For further details, see Appendix~\ref{app:lorasetup}.


For selecting and ordering the in-context examples, we use text-embedding-v3\footnote{\url{https://developer.aliyun.com/article/1567334}} provide by Alibaba for document embedding. 
The vector indexing is implemented with ChromaDB\footnote{\url{https://pypi.org/project/chromadb/}}.
The model names and detail settings are listed in Table~\ref{tab:settings}. \texttt{INSTRUCT} denotes  that additional instructions are used for test instances.


\begin{table}[htbp]
\centering
\small
\setlength\tabcolsep{20pt}
\label{tab:settings}
\begin{tabular}{ccc}
\toprule
Name & Backend Model & Detail Settings \\
\midrule
$M_{\text{LoRA-1}}$ & $M_{\text{LoRA}}$ & $k=20$, \texttt{JOIN} \\
$M_{\text{LoRA-2}}$&$M_{\text{LoRA}}$ & $k=0$, \texttt{JOIN} \\
$M_1$ & Qwen-max & $k=50$, \texttt{SEPA} \\
$M_2$ & Qwen-long & $k=50$, \texttt{SEPA} \\
$M_3$ & Qwen-long & $k=20$, \texttt{JOIN} \\
$M_4$ & Qwen-long & $k=32$, \texttt{JOIN},  \texttt{INSTRUCT} \\
\bottomrule
\end{tabular}
\caption{Different model settings.}
\end{table}


For track-1 and track-2, the ensemble model $M_{\text{E-1}}$ uses $M_1, M_2, M_3, M_4$ with $pos_i=[0.4,0.4,0.2,0.1]$ for track-1 and $pos_i=[0.5,0.4,0.3,0.2]$ for track-2. $\theta_e=-0.5$ for track-1 and  $\theta_e=-0.6$ for track-2, respectively.
We also implement a model $M_{\text{F-1}}$ using the fallback mechanism with $M_1, M_2, M_3, M_4$.
For track-3, the model $M_{\text{F-2}}$ uses the fallback mechanism with $M_{\text{LoRA-1}}, M_{\text{LoRA-2}}, M_3, M_4$.


\begin{table}[t!]
\centering
\setlength\tabcolsep{10pt}
\small
\label{tab:mainresults}
\begin{tabular}{lcccc}
\toprule
Teams & Track-1 & Track-2 & Track-3 & Avg. \\ \midrule
\textbf{ours} & \textbf{47.18} & \textbf{54.03} & \textbf{39.94} & \textbf{47.05} \\ 
BLCU & 43.47 & 51.71 & 38.27 & 44.48 \\ 
YNU & 40.30 & 52.23 & 26.25 & 39.59 \\ 
official baseline & 43.68 & 51.14 & 37.48 & 44.19 \\ \midrule
Single Model-1 (smaller $k$) & 41.88~($M_3$) & 51.05~($M_3$) & 37.07~($M_{\text{LoRA-2}}$) & -- \\ 
Single Model-2 (larger $k$) & 45.23~($M_1$) & 52.52~($M_1$) & 39.39~($M_{\text{LoRA-1}}$) & -- \\ 
Fallback Strategy & 45.24~($M_{\text{F-1}}$) & 52.58~($M_{\text{F-1}}$) & \textbf{39.94}~($M_{\text{F-2}}$) & -- \\ 
Ensemble Strategy & \textbf{47.18}~($M_{\text{E-1}}$) & \textbf{54.03}~($M_{\text{E-1}}$) & -- & -- \\ 
\bottomrule
\end{tabular}
\caption{Results of the official ranking list and some ablation settings. The details of our models are elaborated in Table~\ref{tab:settings} and \S~\ref{subsec:experimentalsetups}.}
\end{table}

\subsection{Results}

The main experimental results are shown in Table~\ref{tab:mainresults}.
Our best method outperforms all other teams and the official baseline on all the three tracks. The average performance is 2.57 higher than the second team.
This result demonstrates the effectiveness of our methods.

The ablation results in Table~\ref{tab:mainresults} are obtained from the official leaderboard\footnote{\url{https://github.com/cubenlp/CERRE-2025CCL/}} as well.
Since the leaderboard only show the performance of the best model per submission (generally, per day), we are unable to report the performances of all models on all tracks.

According to the ablation results, more in-context learning examples (50 v.s. 20) and better backend model (Qwen-max v.s. Qwen-long) bring improvement on track-1 \& track-2 ($M_1>M_3$). 
Interestingly, after LoRA fine-tuning, even the training loss is zero, incorporating 20 in-context learning examples still results in 2.32 improvement  on track-3 ($M_{\text{LoRA-1}}>M_{\text{LoRA-2}}$).
Without ensemble strategies, our singe model (the Single Model-2 line, $M_1~\&~M_{\text{LoRA-1}}$) also outperforms the second-place team on all three tracks by 1.76, 1.81, and 1.12, respectively.

Fallback strategy brings more improvement on track-3 compared to track-1 \& track-2. This is because the open-source LLM (Qwen 2.5 72B) has higher JSON parsing failure rate than the close-source LLM (Qwen-max). Therefore, using the fallback strategy to deal with JSON parsing failure cases is more important.
The ensemble strategy ($M_{\text{E-1}}$) further improves performances on track-1 \& track-2 by 1.94 and 1.45, respectively, obtaining the best results.



\section{Conclusion}
In this paper, we explore how to leverage LLMs to solve the Chinese rhetoric recognition task. 
With both open-source and close-source LLMs, we investigate a series of strategies including LoRA, in-context learning, and model ensemble, and provide empirical results.
We argue that making the JSON outputs more closer to natural language can better leverage LLMs' generation ability, and in-context learning examples still offer benefits even after LoRA fine-tuning.
Our methods achieve an average score of 47.05 on CCL 2025 Chinese essay rhetoric recognition evaluation task, obtaining the first place on all three tracks.
This result demonstrates the effectiveness of our methods.

\section*{Acknowledgements}

This work is supported by NSFC (62206070) and the Innovation Fund Project of the Engineering Research Center of Integration and Application of Digital Learning Technology, Ministry of Education (1421012).


\bibliographystyle{ccl}
\bibliography{anthology}

\appendix

\section{Dataset}
\label{appendix:dataset}

The label sets and statistics of the training data are shown in Table~\ref{table-labelsets} and Table~\ref{trainingdata}, respectively.

\begin{table}[htbp]
\centering
\caption{Label sets for Track-1 and Track-2, and the component list for Track-3.}
\label{table-labelsets}
\begin{tabular}{p{\textwidth}}
\toprule
\textbf{Track-1} : 比喻(明喻, 暗喻, 借喻), 比拟(名词, 动词, 形容词), 夸张(直接夸张, 间接夸张), 排比(成分排比, 句子排比), 反复(间隔反复, 连续反复), 设问(问答连属, 问答不连属), 反问(单句反问, 复句反问), 摹状(通感, 直感)\\
\midrule
\textbf{Track-2} : 比喻(实在物, 动作, 抽象概念)，比拟(拟人, 拟物)，夸张(扩大夸张, 缩小夸张, 超前夸张)，排比(并列, 承接, 递进) \\
\midrule
\textbf{Track-3} : 比喻-明喻（本体，喻体，喻词），比喻-暗喻（本体，喻体），比喻-借喻（喻体），比拟（比拟的特征，比拟的对象），夸张（夸张的对象，夸张的描述），排比（排比词） \\
\bottomrule
\end{tabular}
\end{table}

\begin{table}[htbp]
\centering
\caption{Statistics of the training data}
\label{trainingdata}
\begin{tabular}{p{\textwidth}}
\toprule
\textbf{statistics of the raw training set} :\\
total number of training instances: 50 \\
total number of fine-grained rhetoric devices: 63 (content-level), 105 (form-level)\\
total number of rhetoric component: 145\\
average character number per instance: 98.3\\
average sentence number per instance: 3.7\\
numbers of each rhetoric device in the training set:\\
form-level: 比喻(明喻:20, 暗喻:6, 借喻:6), 比拟(名词:3, 动词:11, 形容词:2), 夸张(直接夸张:7, 间接夸张:2), 排比(成分排比:3, 句子排比:3), 反复(间隔反复:10, 连续反复:1), 设问(问答连属:6, 问答不连属:2), 反问(单句反问:3, 复句反问:1), 摹状(通感:1, 直感:18) \\
content-level: 比喻(实在物:24, 动作:3, 抽象概念:5)，比拟(拟人:13, 拟物:3)，夸张(扩大夸张:8, 缩小夸张:1, 超前夸张:0)，排比(并列:5, 承接:0, 递进:1) \\
\bottomrule
\end{tabular}
\end{table}

\textbf{Evaluation Metrics of Track-1:}
The evaluation score $S_1$ is composed of three parts: F1 score of coarse-grained rhetoric type classification, form-level fine-grained rhetoric type classification, and rhetorical sentence group localization. The specific calculation methods are as follows:

$$
F_1 = 0.3 \times F_1^{\text{rhetoric}} + 0.7 \times F_1^{\text{form}}
$$$$
S_1 = 0.3 \times \text{IoU} + 0.7 \times F_1
$$

Here, $F_1^{\text{rhetoric}}$ and $F_1^{\text{form}}$ represent the $F_1$ scores for coarse-grained and form-level fine-grained rhetorical types, respectively. $\text{IoU}$ denotes the intersection over union value for rhetorical sentence group localization.

\textbf{Evaluation Metrics of Track-2:}
The evaluation score $S_2$ is composed of three parts: F1 score of coarse-grained rhetoric type classification, content-level fine-grained rhetoric type classification, and rhetorical sentence group localization. The specific calculation methods are as follows:

$$
F_1 = 0.3 \times F_1^{\text{rhetoric}} + 0.7 \times F_1^{\text{content}}
$$$$
S_2 = 0.3 \times \text{IoU} + 0.7 \times F_1
$$

Here, $F_1^{\text{rhetoric}}$ and $F_1^{\text{content}}$ represent the $F_1$ scores for coarse-grained and content-level fine-grained rhetorical types, respectively. $\text{IoU}$ denotes the intersection over union value for rhetorical sentence group localization.

\textbf{Evaluation Metrics of Track-3:}
The evaluation score $S_3$ is composed of four parts: the F1 scores of conjunctions, tenors, vehicles, and rhetorical sentence group localization. The specific calculation methods are as follows:

$$
F_1 = \frac{1}{3} \times F_1^{\text{conjunction}} + \frac{1}{3} \times F_1^{\text{tenor}} + \frac{1}{3} \times F_1^{\text{vehicle}}
$$$$
S_3 = 0.3 \times \text{IoU} + 0.7 \times F_1
$$

Here, $F_1^{\text{conjunction}}$, $F_1^{\text{tenor}}$, and $F_1^{\text{vehicle}}$ represent the $F_1$ scores for conjunctions, tenors, and vehicles, respectively. $\text{IoU}$ denotes the intersection over union value for rhetorical sentence group localization.

\section{System Prompt}
\label{appendix:system_prompt}

The system prompt is shown below, which is used in both LoRA and in-context learning.
Within the system prompt, we provide LLMs with clear definitions of rhetorical types and components, accompanied by illustrative examples. 

你是一名经验丰富的小学语文老师，在研究中国小学生作文修辞的特点。 \\
以下是8种修辞类型定义，形式上与内容上的细粒度分类，以及案例。你对此都十分熟悉，并能够灵活运用。

\begin{enumerate}
\item 比喻 \\
比喻是通过将一事物比作另一事物，突出其特点或共同属性的修辞手法。 \\
形式上的细粒度分类：
\begin{itemize}
    \item 明喻: 明确表达本体、喻体和比较关系。 \\
      案例: "庄稼汉们站在地头，望着这片黄澄澄像狗尾巴的稻谷，心里像睡了蜜一样的甜" — 本体是稻谷，喻体是狗尾巴，喻词是“像”。
    \item 暗喻: 不直接表达但暗示本体与喻体的关系。 \\
      案例: "时间就是金钱" — 本体是时间，喻体是金钱；此处喻词暗含在“就是”中。
    \item 借喻: 只提及喻体，通过上下文让读者推断本体。 \\
      案例: "一把弯刀挂在天上" — 本体是月亮（文中没有提及，通过上下文推测），喻体是弯刀；借喻通常不明显使用喻词，而是依靠语境引导。
\end{itemize}
内容上的细粒度分类：
\begin{itemize}
    \item 实在物: 涉及具体的物体或现象。 \\
      案例: "一把弯刀挂在天上" — 将月亮比作弯道，属于实在物。
    \item 动作: 涉及具体的动作、行为或事件。 \\
      案例: "丁尽烟华的一夜雨声，敲起了春耕的锄，播响了播种的鼓" — 本体是雨声，喻体是敲起了春耕的锄，播响了播种的鼓，属于动作。
    \item 抽象概念: 涉及不可直接感知的抽象概念。 \\
      案例: "时间就是金钱" — 本体是时间，喻体是金钱，属于抽象概念。
\end{itemize}

\item 比拟 \\
比拟是通过赋予非人事物以人的特征或行为或赋予某一事物以其他事物的特征或行为来辅助描述的一种修辞手法。 \\
形式上的细粒度分类：
\begin{itemize}
    \item 名词： \\
      案例：我的这辆车久历风尘，实在高寿。 - 比拟对象：车 - 比拟特征：通常描述人的名词（高寿）
    \item 动词： \\
      案例：我到我家的房外，我的母亲早已迎着出来了，接着便飞出了几岁的宠儿法儿。 --  比拟对象：宠儿法儿的行为   -  比拟特征：物品飞出的行为
    \item 形容词： \\
      案例：湖水忽发温柔，忽发安详。 -   比拟对象：湖水  -  比拟特征：使用描述人的形容词（温柔、安详）
\end{itemize}
内容上的细粒度分类：
\begin{itemize}
    \item 拟人： \\
      案例：湖水忽发温柔，忽发安详。  -  比拟对象：湖水  -  比拟特征：温柔、安详，属于人类的情感
    \item 拟物： \\
      案例：我到我家的房外，我的母亲早已迎着出来了，接着便飞出了几岁的宠儿法儿。 -  比拟对象：宠儿法儿的行为  -  比拟特征：人是不会飞出的，而飞出的是物，所以属于拟物
\end{itemize}

\item 夸张 \\
夸张是通过极端放大或缩小事物的特性来强调其效果。 \\
形式上的细粒度分类：
\begin{itemize}
    \item 直接夸张：直接对事物的某个方面进行夸张描述。 \\
      案例："而正前方，是那块巴掌大的小绿茵地和两栋小教学楼。" - 夸张对象：小绿茵地和两栋小教学楼 - 夸张的描述：巴掌大
    \item 间接夸张：通过夸大与主题相关的其他事物来间接强调主题。 \\
      案例："小明激动地跳来跳去，眼睛亮得仿佛可以放射出火花。"  - 夸张对象：小明激动地跳来跳去 - 夸张的描述：眼睛亮得仿佛可以放射出火花
\end{itemize}
内容上的细粒度分类：
\begin{itemize}
    \item 扩大夸张：向大、多、长或高等方向夸大。 \\
      案例："那一刻，力量再次回升，我进入最后冲刺，一下超了三个同学，跑道两旁已是虚景。" - 夸张对象：我 - 夸张的描述：跑道两旁已是虚景，终点是蓝蓝的
    \item 缩小夸张：向小、少、短或低等方向夸大。 \\
      案例："而正前方，是那块巴掌大的小绿茵地和两栋小教学楼。" - 夸张对象：小绿茵地和两栋小教学楼 - 夸张的描述：巴掌大
    \item 超前夸张：把后出现的事说到先出现的事之前 \\
      案例：他抓了一辈子鱼，吃了一辈子鱼，却从没感到过腻。   - 夸张对象：吃了一辈子鱼、抓了一辈子鱼  - 夸张的描述：一辈子、一辈子
\end{itemize}

\item 排比 \\
排比使用重复或并列的结构来增强语言的节奏感和表达力。 \\
形式上的细粒度分类：
\begin{itemize}
    \item 成分排比：排比项作为句子中的某个成分出现，如主语或宾语。 \\
      案例：“这启示了我们时机是因人而异的，有的人年少成名，有的人在中年创下辉煌，有的人大器晚成。”排比引导词为“有的”，排比项不能独立构成句子。
    \item 句子排比：每个排比项都是一个完整的句子，独立且表达完整的意义。 \\
      案例：“我想，总有一些穿板鞋走不到的路，总有一些柏油马路上闻不到的空气，总有一些在高楼大厦里遇不见的人。” 排比引导词为“总有”，每个排比项都是一个完整的句子。
\end{itemize}
内容上的细粒度分类：
\begin{itemize}
    \item 并列：排比项之间并列，改变顺序不影响整体意义。 \\
      案例：“我想，总有一些穿板鞋走不到的路，总有一些柏油马路上闻不到的空气，总有一些在高楼大厦里遇不见的人。” 排比引导词为“总有”，排比项之间顺序无关
    \item 承接：排比项之间有逻辑顺序，顺序改变会影响意义。 \\
      案例：“这启示了我们时机是因人而异的，有的人年少成名，有的人在中年创下辉煌，有的人大器晚成。”排比引导词为“有的”，排比项之间有年龄的顺序承接关系。
    \item 递进：排比项之间按照程度、时间或其他因素逐渐升级或加强。 \\
      案例：“读书，读好书，读完整的书，爱书，此为读者。”，排比引导词为“读”，引导了一种语气渐强的递进关系。
\end{itemize}

\item 反复 \\
反复是通过重复特定词语或句子结构来加强语气、情感表达的一种修辞手法。 \\
形式上的细粒度分类：
\begin{itemize}
    \item 间隔反复: 提挈语不紧密贴连。 \\
      案例: "哪里是山，哪里是房屋，哪里是菜园，我终于分辨出来了。" — 挈语为“哪里是”。
    \item 连续反复: 提挈语连续出现。 \\
      案例: 还拿天气说吧，老那么好，老那么好，没有变化，没有春夏秋冬，这就使人生厌。 — 挈语为“老那么好”。
\end{itemize}

\item 设问 \\
设问是在文本中主动提出问题，随后自行给出答案，以引发读者思考和关注的修辞手法。 \\
形式上的细粒度分类
\begin{itemize}
    \item 问答连属: 问和答连在一起。 \\
      案例: 这是我的决定吗？是的。  -  问题：“这是我的决定吗？”，回答：“是的。”
    \item 问答不连属: 问和答没有连在一起。 \\
      案例: 理论和事实比较起来，哪一个更重要呢？这个问话好像是多余的。因为理论是理性知识，对事实的...... -  问题：“哪一个更重要呢？”，回答：”因为理论是理性知识，对事实的“。
\end{itemize}

\item 反问 \\
反问是通过不期待回答的形式提出问题，以强化语气和表达某种观点的修辞手法。 \\
形式上的细粒度分类：
\begin{itemize}
    \item 单句反问: 反问句所属句子是单句。 \\
      案例: 难道你不明白这个道理吗？ - 反问句是单句
    \item 复句反问: 反问句所属句子是复句。 \\
      案例: 如果努力都没有意义，那么世界上还有什么可以期待的呢？  - 反问句是复句，还包含“如果”那半句
\end{itemize}

\item 摹状 \\
摹状是通过对事物的具体形态、特征进行形象化描绘，以增强语言表现力的一种修辞手法。 \\
形式上的细粒度分类：
\begin{itemize}
    \item 通感: 摹状词以及所描述的状不同，A摹状词描述B状。 \\
      案例：秋风里，时时有玉钱蝴蝶，翩翩飞来，停在花上，好半天不动，幽情凄恋。它要僵了，它愿意僵在花儿的冷香里！- 冷香是嗅觉，却来通感“要僵了”的温度
    \item 直感: 摹状词以及所描述的状同类，A摹状词描述A状。 \\
      案例：两把芭蕉扇做的脚踏车，麻雀牌堆成的火车，汽车，你何等认真地看待，挺直了嗓子叫“汪——”，“咕咕咕……”，来代替汽笛。 - 叫声和汽笛声都是声音。
\end{itemize}

\item 没有修辞 \\
没有以上8种类型的修辞手法。
\end{enumerate}

\section{In-context Learning Prompt}
\label{app:iclp}

An illustration of in-context learning prompt is shown bellow. Here, we assume that:

${sim}(\texttt{INPUT DOCUMENT$_1$},$ $ \texttt{INPUT DOCUMENT$_{\texttt{TEST}}$} )$ $ \leq {sim}(\texttt{INPUT DOCUMENT$_2$}, $ $\texttt{INPUT DOCUMENT$_{\texttt{TEST}}$} )$ $\leq \ldots$ $ \leq {sim}(\texttt{INPUT DOCUMENT$_k$}, $ $\texttt{INPUT DOCUMENT$_{\texttt{TEST}}$} )$

\noindent where $sim(\dot)$ is the text similarity function.

\begin{table}[htbp]
\centering
\begin{tabular}{p{\textwidth}}
[\{``role": ``system", ``content": \texttt{SYSTEM PROMPT}\}, \{``role": ``user", ``content": \texttt{INPUT DOCUMENT$_1$}\}, \{``role": ``assistant", ``content": \texttt{JSON OUTPUT$_1$}\}, \{``role": ``user", ``content": \texttt{INPUT DOCUMENT$_2$}\}, \{``role": ``assistant", ``content": \texttt{JSON OUTPUT$_2$}\}, ......, \{``role": ``user", ``content": \texttt{INPUT DOCUMENT$_k$}\}, \{``role": ``assistant", ``content": \texttt{JSON OUTPUT$_k$}\}, \{``role": ``user", ``content": \texttt{INPUT DOCUMENT$_{\texttt{TEST}}$}\}] \\
\end{tabular}
\end{table}

\section{LoRA Setup Details}
\label{app:lorasetup}
We use Qwen 2.5 72B \cite{qwen2.5} as backend for LoRA. The model is trained for 24 epochs with batch size of 16, i.e., the training has 75 steps in total. A linear learning rate scheduler is utilized with peak learning rate of 3e-4. The 5\% of the total training steps are used for warm-up. Weight decay is 0.01. For LoRA-specific configurations, the rank value is 8, the alpha scaling factor is 32, and the dropout rate is 0.1. All eligible layers within the model are applied LoRA method.
We implement the method with Alibaba cloud model studio.


\section{Error Analysis}
Because only 50 instances exist in the training set, without annotated validation data, it is unrealistic to obtain statistically meaningful misclassifications analysis. 
We calculate the error rates of no output errors (due to LLMs' content safety restrictions), JSON parsing errors in Table~\ref{tab:app:error_1}.
We also show the numbers of rhetoric predictions where the coarse/fine-grained rhetoric types are not in the corresponding label sets.

We can see that the error rates are much high for $M_1$ (Joint Training  Part) than $M_1$ (Trace-1-only Part), we think it is because in the latter setting, the task is easier and the prompt is more concise.
Comparing $M_{\text{LoRA-1}}$ and $M_{\text{LoRA-2}}$, we find that the no output rate and the JSON parsing error rate is much lower for $M_{\text{LoRA-2}}$. Without in-context learning examples, the data format in inference is more consistent to that during training, which improves the success rate of generated output formats. However, without in-context examples, $M_{\text{LoRA-2}}$ tends to generates more rhetoric types that are not exists in the label sets, thus degenerate the performances.

\begin{table}[htbp]
\centering
\caption{Statistics of inference error rates. The details of our models are elaborated in Table~\ref{tab:settings} and \S~\ref{subsec:experimentalsetups}.}
\label{tab:app:error_1}
\begin{tabular}{p{6cm}p{2cm}p{2cm}p{2cm}p{2cm}}
\toprule
model name & No Output (\%) & JSON Parsing Error (\%) & \# Wrong Coarse-grained Type & \# Fine-grained Type \\ \midrule
$M_{\text{LoRA-1}}$ & 0.15 & 1.20 & 428 & 5  \\
$M_{\text{LoRA-2}}$ & 0.05 & 0.47 & 1024 & 4 \\
$M_1$ (Joint Training Part. Rhetoric types: metaphor, personification, hyperbole, and parallelism) & 0.07 & 1.19 & 358 & 3 \\
$M_1$ (Trace-1-only Part. Rhetoric types:  rhetorical question, hypophora, repetition, synesthesia) & 0.01 & 0.07 & 1 & 0 \\
\bottomrule
\end{tabular}
\end{table}

We further demonstrate some JSON parsing error cases in Table~\ref{app:tab:jsonerror}.

\begin{table}[htbp]
\centering
\caption{Cost for model training and inference (CNY). $M_{\text{LoRA}}$ is the Qwen 2.5 72B model after LoRA continue-training. $M_{\text{LoRA-2}}$ is the inference model with $M_{\text{LoRA}}$, joint setting with all rhetoric types, without in-context learning example. $M_{\text{LoRA-1}}$ is similar to $M_{\text{LoRA-2}}$, but incorporating the in-context learning strategy, with in-context example number $k=20$ and similarity-based example selection and ordering strategies. $M_1$ is an in-context learning based inference model, with Qwen-max as the backend model, in-context example number $k=50$, and the similarity-based example ordering strategy, focusing on the joint training part (see \S~\ref{subsec:example}). }
\label{cost}
\setlength\tabcolsep{3pt}
\begin{tabular}{p{8cm}ccc}
\toprule
model name & train cost & inference cost & price per million tokens\\ \midrule
$M_{\text{LoRA}}$ & 460.36 & -- & 150 \\
$M_{\text{LoRA-1}}$ & -- & 1187.66  & 4(input), 12(output)\\
$M_{\text{LoRA-2}}$ & -- & 430.50 & 4(input), 12(output)\\
$M_1$ (Joint Training Part. Rhetoric types: metaphor, personification, hyperbole, and parallelism) & -- & 1083.42  & 2.4(input), 9.6(output)\\
\bottomrule
\end{tabular}
\end{table}

\begin{table}[htbp]
\centering
\caption{Case study for JSON parsing error.}
\label{app:tab:jsonerror}
\begin{tabular}{p{\textwidth}}
\toprule
\textbf{model }: $M_{\text{LoRA-1}}$ \\
\textbf{content}: "[{"内容片段": "上次还没被妈妈骂够吗？”, "修辞手法": "反问", "形式上的细粒度修辞分类": "单句反问"}]\\
\textbf{error type}: erroneous predicts " (\textbackslash u0022) as ” (\textbackslash u201d) . \\\midrule
\textbf{model }: $M_{\text{LoRA-1}}$ \\
\textbf{content}:  [{"内容片段": "我的家，是一个“小小动物园"里面有四个动物，分别是：“黄牛妈妈、大棕能爸爸、猴子弟弟和小鱼的我。", "修辞手法": "比喻", "形式上的细粒度修辞分类": "暗喻", "内容上的细粒度修辞分类": "实在物", "本体": "我的家", "喻体": "一个“小小动物园""}, {"内容片段": "我的家，是一个“小小动物园"里面有四个动物，分别是：“黄牛妈妈、大棕能爸爸、猴子弟弟和小鱼的我。", "修辞手法": "排比", "形式上的细粒度修辞分类": "成分排比", "内容上的细粒度修辞分类": "并列", "排比词": "动物"}, {"内容片段": "说起我的妈妈，她肯定是一个不折不扣的＂老黄牛"啦。", "修辞手法": "比喻", "形式上的细粒度修辞分类": "暗喻", "内容上的细粒度修辞分类": "实在物", "本体": "我的妈妈", "喻体": "一个不折不扣的＂老黄牛""}]\\
\textbf{error type}: multiple errors in generating quotation marks.\\ \midrule
\textbf{model }: $M_1$ (Joint Training Part. Rhetoric types: metaphor, personification, hyperbole, and parallelism)  \\
\textbf{content}: 这段文字中有一些语法和拼写错误，但还是可以从中找到一些修辞手法。以下是其中的修辞分析：\\
1. \textbf{比喻}：
   \begin{itemize}
     \item ``在网上是玩的了在那里了玩具，和一些大儿玩手工和无人机和大儿和小儿玩和教古箱和恐龙等都是可以玩的在水里间很有玩了''：这里虽然表达有些混乱，但``在水里间很有玩了''可能是在形容水中游乐设施的乐趣。
     \item ``比喻星球大作蛇你要喝水你可以在那里卖可口可乐很好喝是无塘的''：这里的``比喻星球大作蛇''可能是想说``比如《星球大战》''。
   \end{itemize}
   2. \textbf{比拟}：
   \begin{itemize}
     \item ``小蜘蛛开始时有四个钢条可以让利间让人就杀了，大车时小蜘蛛就开始了这个巨就杀了一百多个但最后力宝石一飞''：这里的``小蜘蛛''被赋予了人的行为，如``开始时有四个钢条可以让利间让人就杀了''，这是拟人化的表达。
   \end{itemize}
   3. \textbf{夸张}：
   \begin{itemize}
     \item ``小蜘蛛开始时有四个钢条可以让利间让人就杀了，大车时小蜘蛛就开始了这个巨就杀了一百多个但最后力宝石一飞''：这里的``小蜘蛛''能够``杀了一百多个''是一种夸张的说法，用来强调游戏的激烈程度。
   \end{itemize}
   总结起来，这段文字中使用了比喻、比拟和夸张等修辞手法，但由于表达较为混乱，部分句子需要进一步澄清和修正。以下是修正后的版本：\\
   ---
有一天，大家好，我是小张。今天我要去苏州湾东一日游。你们去过那里吗？可以说一说哦。\\
我们坐着地铁，十几分钟后经过了5个站点，来到了苏州湾。从1号入口进入第一层，以前这里是正方形的，现在变成了圆形，电梯在中间。坐上电梯，可以看到很多卖汽车的地方。现在第一层又变成了正方形，在网上有很多好玩的东西，比如玩具、手工制作、无人机等。大人和小孩都可以在这里玩，还有一些古装体验和恐龙模型等，非常有趣。\\
中午吃饭的时候，你可以在这里买到可口可乐，非常好喝而且是无糖的。下午我们可以去看电影，比如《星球大战》。还有一个地方可以玩VR游戏，里面有很多种不同的游戏，比如小蜘蛛游戏，一开始有四个钢条可以让玩家通过，然后小蜘蛛开始行动，这个游戏非常刺激，最后还有宝石飞出来。\\
今天就是苏州湾的一日游，如果你去了什么好玩的地方，一定要告诉我哦。再见！\\
---
这句话似乎有一些语法和拼写错误，我将其修正为：``我们依依不舍地离开了秀峰，回到了我的外公家。''\\
修正后的句子没有使用任何修辞手法。如果有其他内容或需要进一步分析，请告诉我！ \\
\textbf{error type}: JSON format collapse.\\
\bottomrule
\end{tabular}
\end{table}

\section{Cost Analysis}

We report the cost for model training and inference in CNY in Table~\ref{cost}.
$M_{\text{LoRA}}$ is the model after LoRA fine-tuning, serving as the backend model for $M_{\text{LoRA-2}}$ \& $M_{\text{LoRA-1}}$.
For $M_1$, we only report the inference cost for the joint training part, which is the most expensive part of $M_1$.

\end{CJK*}
\end{document}